\documentclass[10pt,twocolumn,letterpaper]{article}

\usepackage[pagenumbers]{cvpr} 

\usepackage{graphicx}
\usepackage{amsmath}
\usepackage{amssymb}
\usepackage{booktabs}
\usepackage{enumitem}

\usepackage{cuted}
\usepackage{multirow}

\usepackage[pagebackref,breaklinks,colorlinks]{hyperref}

\usepackage[capitalize]{cleveref}
\crefname{section}{Sec.}{Secs.}
\Crefname{section}{Section}{Sections}
\Crefname{table}{Table}{Tables}
\crefname{table}{Tab.}{Tabs.}

\def\cvprPaperID{4460} 
\def\confName{CVPR}
\def\confYear{2023}

\begin{document}

\newcommand{\reviseme}[1]{\textcolor{red}{#1}}

\newcommand{\xiaoshuai}[1]{\textcolor{magenta}{[Xiaoshuai:#1]}}
\newcommand{\kyle}[1]{\textcolor{blue}{[Kyle:#1]}}
\newcommand{\tom}[1]{\textcolor{green}{[Tom:#1]}}


\title{Nerflets: Local Radiance Fields for Efficient Structure-Aware\\
3D Scene Representation from 2D Supervision}

\author{
Xiaoshuai Zhang$^{1,3}$
\,\,
Abhijit Kundu$^1$
\,\, 
Thomas Funkhouser$^1$ \\
Leonidas Guibas$^{1,2}$
\,\, 
Hao Su$^3$
\,\,
Kyle Genova$^1$\\
{\small $^{1}$ Google Research \qquad $^{2}$ Stanford University  \qquad $^{3}$ University of California, San Diego}}
\maketitle

\newcommand{\bfr}{\mathbf{r}}
\newcommand{\bfx}{\mathbf{x}}
\newcommand{\bfo}{\mathbf{o}}
\newcommand{\bfd}{\mathbf{d}}
\newcommand{\bfc}{\mathbf{c}}
\newcommand{\bfw}{\mathbf{w}}
\newcommand{\bfs}{\mathbf{s}}
\newcommand{\bfmu}{\boldsymbol{\mu}}

\newcommand{\calL}{\mathcal{L}}

\setlength{\floatsep}{0.50\floatsep}
\setlength{\textfloatsep}{0.50\textfloatsep}
\setlength{\dblfloatsep}{0.50\dblfloatsep}
\setlength{\dbltextfloatsep}{0.50\dbltextfloatsep}
\setlength{\abovedisplayskip}{3pt}
\setlength{\belowdisplayskip}{3pt}

\begin{strip}
  \centering
  \vspace{-4.3em}
  \includegraphics[width=\textwidth]{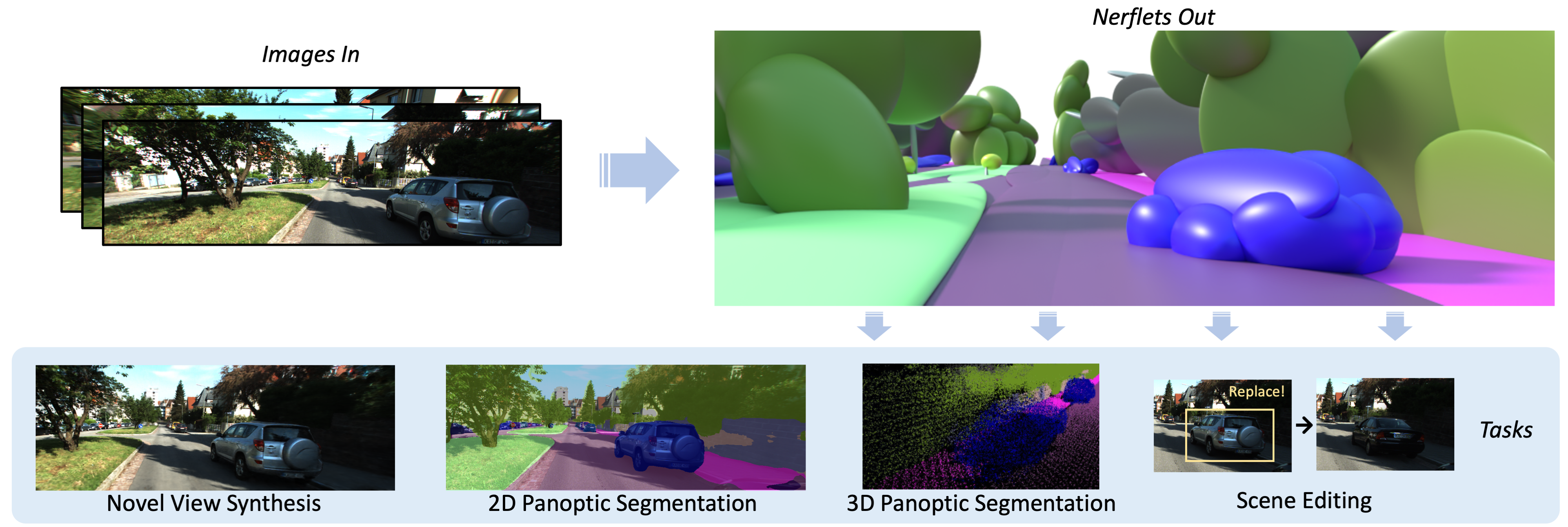}
  \vspace{-1.8em}
  \captionof{figure}{We propose to represent the scene with a set of local neural radiance fields, named nerflets, which are trained with only 2D supervision. Our representation is not only useful for 2D tasks such as novel view synthesis and panoptic segmentation, but also capable of solving 3D-oriented tasks such as 3D segmentation and scene editing. The key idea is our learned structured decomposition (top right).}
  \vspace{-0.4em}
  \label{fig:teaser}
\end{strip}

\begin{abstract}
\vspace{-1em}

We address efficient and structure-aware 3D scene representation from images. \emph{Nerflets} are our key contribution-- a set of local neural radiance fields that together represent a scene. Each nerflet maintains its own spatial position, orientation, and extent, within which it contributes to panoptic, density, and radiance reconstructions. By leveraging only photometric and inferred panoptic image supervision, we can directly and jointly optimize the parameters of a set of nerflets so as to form a decomposed representation of the scene, where each object instance is represented by a group of nerflets.   During experiments with indoor and outdoor environments, we find that nerflets: (1) fit and approximate the scene more efficiently than traditional global NeRFs, (2) allow the extraction of panoptic and photometric renderings from arbitrary views, and (3) enable tasks rare for NeRFs, such as 3D panoptic segmentation and interactive editing. \href{https://jetd1.github.io/nerflets-web/}{Our project page}.

\end{abstract}
\vspace{-2em}

\section{Introduction}
\label{sec:intro}

This paper aims to produce a compact, efficient, and comprehensive 3D scene representation from only 2D images. Ideally, the representation should reconstruct appearances, infer semantics, and separate object instances, so that it can be used in a variety of computer vision and robotics tasks, including 2D and 3D panoptic segmentation, interactive scene editing, and novel view synthesis.





Many previous approaches have attempted to generate rich 3D scene representations from images. PanopticFusion~\cite{narita2019panopticfusion} produces 3D panoptic labels from images, though it requires input depth measurements from specialized sensors. NeRF~\cite{mildenhall2021nerf} and its descendants~\cite{mueller2022instant, barron2021mip, barron2022mipnerf360, reiser2021kilonerf} produce 3D density and radiance fields that are useful for novel view synthesis, surface reconstruction, semantic segmentation~\cite{Zhi21iccv_SemanticNeRF, vora2021nesf}, and panoptic segmentation~\cite{kundu2022panoptic, wang2022dmnerf}. However, existing approaches require 3D ground truth supervision, are inefficient, or do not handle object instances.



We propose \emph{nerflets}, a 3D scene representation with multiple local neural fields that are optimized jointly to describe the appearance, density, semantics, and object instances in a scene (Figure~\ref{fig:teaser}). Nerflets constitute a \textit{structured} and \textit{irregular} representation-- each is parameterized by a 3D center, a 3D XYZ rotation, and 3 (per-axis) radii in a 9-DOF coordinate frame.
The influence of every nerflet is modulated by a radial basis function (RBF) which falls off with increasing distance from the nerflet center according to its orientation and radii, ensuring that each nerflet contributes to a local part of the scene.  Within that region of influence, each nerflet has a miniature MLP to estimate density and radiance. It also stores one semantic logit vector describing the category (e.g., ``car'') of the nerflet, and one instance label indicating which real-world object it belongs to (e.g., ``the third car''). In Figure~\ref{fig:teaser}, each ellipsoid is a single nerflet, and they are colored according to their semantics.

A scene can contain any number of nerflets, they may be placed anywhere in space, and they may overlap, which provides the flexibility to model complex, sparse 3D scenes efficiently.  Since multiple nerflets can have the same instance label, they can combine to represent the density and radiance distributions of complex object instances.   Conversely, since each nerflet has only one instance label, the nerflets provide a complete decomposition of the scene into real-world objects. Nerflets therefore provide a 3D panoptic decomposition of a scene that can be rendered and edited.

Synthesizing images using nerflets proceeds with density-based volume rendering just as in NeRF~\cite{mildenhall2021nerf}.  However, instead of evaluating one large MLP at each point sample along a ray, we evaluate the small MLPs of only the nerflets near a sample. We average the results, weighting by the influence each nerflet has over that sample.  The rendering is fully-differentiable with respect to all continuous parameters of the nerflets. Fitting the nerflet representation is performed from a set of posed RGB images with a single training stage. After training, instance labels are assigned based on the scene structure, completing the representation.

Experiments with indoor and outdoor datasets confirm the main benefits of nerflets.
We find that: 
1) Parsimony encourages the optimizer to decompose the scene into nerflets 
with consistent projections into novel panoptic images (Section~\ref{sec:exp_panop2d});
2) Semantic supervision can be beneficial to novel view synthesis (Section~\ref{subsec:semvsapp});
3) Structure encourages efficiency, compactness, and scalability (Section~\ref{sec:effient_evaluation}); and
4) the explicit decomposition of a scene improves human interpretability for easy interactive editing, including adding and removing objects (Section~\ref{subsec:exp_editing}).
These benefits enable state-of-the-art performance on the KITTI360~\cite{kitti360} novel semantic view synthesis benchmark, competitive performance on ScanNet 3D panoptic segmentation tasks with more limited supervision, and an interactive 3D editing tool that leverages the efficiency and 3D decomposition of nerflets.   

The following summarizes our main contributions:
\begin{itemize}[itemsep=0.1pt,topsep=2pt,leftmargin=2em]
    \item We propose a novel 3D scene representation made of small, posed, local neural fields named \emph{nerflets}.
    \item The pose, shape, panoptic, and appearance information of nerflets are all fit jointly in a single training stage, resulting in a comprehensive learned 3D decomposition from real RGB images of indoor or outdoor scenes.
    \item 
    We test nerflets on 4 tasks- novel view synthesis, panoptic view synthesis, 3D panoptic segmentation and reconstruction, and interactive editing.
    \item We achieve 1st place on the KITTI-360 semantic novel view synthesis leaderboard.
\end{itemize}

\section{Related Work}
\label{sec:rel}
Recently, the success of deep learning approaches for both computer vision and graphics tasks has enabled researchers to reconstruct and reason about 3D scenes under various settings. We review related work on segmentation and neural field based scene representations.

\vspace*{2mm}\noindent
\textbf{Semantic, Instance, and Panoptic Segmentation:} There are many methods designed for semantic, instance, and/or panoptic~\cite{kirillov2019panoptic} segmentation.
The most popular approaches are fully-supervised and operate within a single input data modality.
For example, 2D approaches~\cite{ronneberger2015u,long2015fully,badrinarayanan2017segnet,chen2017deeplab,zhao2017pyramid,zheng2021rethinking,he2017mask,liu2021swin,zhang2022resnest} are usually based on CNN or transformer backbones and associate each pixel in an image with certain semantic or instance labels. We leverage a trained 2D panoptic model, Panoptic Deeplab~\cite{cheng2020panoptic}, in our framework.

Similar frameworks have been proposed to solve 3D segmentation tasks for 3D point clouds \cite{pham2019jsis3d,qi2017pointnet,qi2017pointnet++,shi2019pv,thomas2019kpconv}, meshes \cite{hanocka2019meshcnn,huang2019texturenet}, voxel grids \cite{graham2015sparse,song2017semantic}, and octrees \cite{riegler2017octnet}.
However, these methods typically require a large amount of annotated 3D data, which is expensive to obtain.

To avoid the need for 3D annotations, several multiview fusion approaches have explored aggregating 2D image semantic features onto a pointcloud or mesh using
weighted averaging \cite{hermans2014dense,kundu2020virtual,vineet2015incremental, armeni20193d}, conditional random fields (CRFs) \cite{kundu2014joint,mccormac2017semanticfusion}, and bayesian fusions \cite{ma2017multi,vineet2015incremental,zhang2019large}. There have also been approaches like 2D3DNet~\cite{genova2021learning} that combine both 2D mutiview fusion with a 3D model.

In contrast to these methods, ours builds a complete 3D representation including geometry, appearance, semantic, and instance information from only 2D inputs and without any input 3D substrate such as a mesh or a point cloud.

\vspace*{2mm}\noindent
\textbf{Scene Understanding with NeRF:}
NeRF~\cite{mildenhall2021nerf} and subsequent work~\cite{kundu2022panoptic,ost2021neural,yang2021objectnerf,Zhi:etal:ICCV2021,vora2021nesf,kobayashi2022decomposing} show the promise of neural radiance fields for tasks beyond novel view synthesis, including 3D reconstruction, semantic segmentation, and panoptic segmentation.   
For example, SemanticNeRF~\cite{Zhi:etal:ICCV2021}, and NeSF~\cite{vora2021nesf} are useful for semantic understanding, but do not consider object instances.
DFF~\cite{kobayashi2022decomposing} leverages the power of large language models for semantic understanding, but similarly does not produce object instances.
ObjectNeRF~\cite{yang2021objectnerf} and NSG~\cite{ost2021neural} are useful for object editing, but do not produce a full panoptic decomposition or support efficient interactive editing.  None of these methods produce a complete scene representation in the way that nerflets do.

Panoptic Neural Fields (PNF)~\cite{kundu2022panoptic} is very relevant to our paper as it supports both semantic scene understanding and object-level scene editing.  PNF first runs a 3D object detector and then a tracker to create an input set of object tracks.   They then fit an individual MLP for each object track and another special ``stuff'' MLP for the remainder of the scene. 
This is a compelling and effective approach which supports moving objects, but it does not solve our target problem. It 1) requires expensive ground truth 3D supervision for the detector and tracker that are only available for some classes, and
2) has a fixed 3D scene decomposition that is provided by the input tracker results.
This last point means that it can fail when the detector or tracker fails, even if a multi-view analysis-by-synthesis appearance loss, like in NeRF, would have been able to force a correct prediction by requiring all pixels to be described by some object instance.
By comparison, nerflets require only 2D supervision, support any class for which 2D panoptic segmentations are available, and optimize most parameters jointly, improving efficiency and instance recall.

DM-NeRF~\cite{wang2022dmnerf} is highly related concurrent work. It learns an object decomposition of a scene, but does not provide the explicit structure, full panoptic decomposition, or easy interactive editing of nerflets. In particular, a large MLP decodes spatial positions to object identity vectors, and editing therefore requires careful consideration-- an inverse query algorithm~\cite{wang2022dmnerf}. By contrast, nerflets can be edited directly as geometric primitives. We compare quantitatively to both PNF and DM-NeRF in Section~\ref{sec:exp}.

\vspace*{2mm}\noindent
\textbf{Structured NeRF Representations:} 
One of the key advantages of nerflets is their irregular structure, which has been investigated in other contexts. Many existing approaches exploit structure for efficiency~\cite{mueller2022instant,reiser2021kilonerf,SunSC22,wu2021diver,lombardi2021mixture}, compactness~\cite{liu2020neural}, scalability~\cite{turki2022mega,zhang2022nerfusion}, human-interpretability~\cite{ost2021neural}, parsimony~\cite{genova2020local}, or editability\cite{kundu2022panoptic,yang2021objectnerf}.
For example, Kilo-NeRF~\cite{reiser2021kilonerf} and DiVER~\cite{wu2021diver} exploit regular grids of MLPs or features to improve the efficiency of NeRF novel view synthesis. MVP~\cite{lombardi2021mixture} builds an irregular primitive-based representation for real-time portrait rendering, but requires explicit scene geometry inputs to initialize the primitives and freezes their location after initialization. We take inspiration from these approaches, which achieve impressive performance through local structure, and apply their insights to panoptic segmentation and editing. Unlike these approaches, nerflets can conform to an object's extent and then move as it is edited.
In the future, more benefits of exploring irregular NeRF representations could include tracking a moving object or allowing for a consistent local coordinate frame for learning 3D priors.



\section{Method}
\label{sec:method}
This section introduces our nerflet scene representation and our training and rendering method. As in NeRF~\cite{mildenhall2021nerf}, the input to our method is a set of posed 2D RGB images.
We first run an off-the-shelf 2D panoptic segmentation model~\cite{cheng2020panoptic} to generate predicted 2D semantic and instance images, which we use as a target during the optimization. 
Next, we optimize our core nerflet representation (Section~\ref{representation}) to convergence on photometric, semantic, instance, and regularization losses applied to images rendered with volumetric ray-marching~\cite{mildenhall2021nerf} (Section~\ref{sec:loss_functions}).
Finally, we assign instance labels to the nerflets based on the learned decomposition (Section~\ref{sec:instance_assignment}), at which point the representation is complete and ready for rendering or editing.



\subsection{Scene Representation}
\label{representation}

The core novelty of our framework is the nerflet scene representation. Nerflets are a structured representation, where the output radiance and panoptic fields are defined by blending the values produced by $N$ individual nerflets. 



\noindent\textbf{Nerflet definition:} 
\begin{figure}[t]
  \centering
  \includegraphics[width=1.0\linewidth]{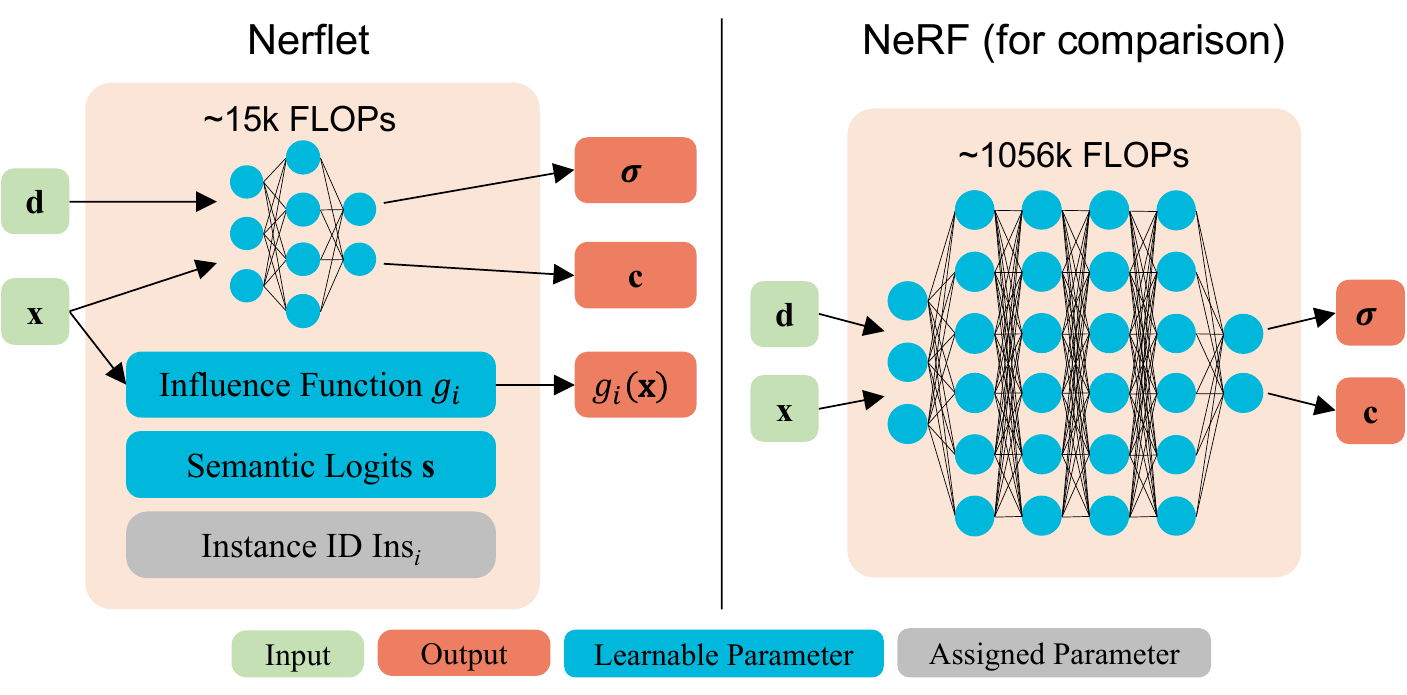}
   \caption{Information maintained by a nerflet and NeRF. Compared to NeRF, a nerflet focuses only on a small portion of the scene determined by its influence function $g$ (Eq.~\ref{eq:influence_fun}), and thus it uses a miniature MLP to fit density $\sigma$ and color $\bfc$. Each nerflet also maintains a single semantic logit vector $\bfs_i$ and an assigned instance ID $\text{Ins}_i$. Together these parameters comprise a compact building block for our scene representation.}
   \label{fig:nerflet_info}
\end{figure}
Each nerflet stores local geometry, appearance, semantic, and instance information. As shown in Figure.~\ref{fig:nerflet_info}, a nerflet $i$ has 1) position and orientation parameters that define its influence function $g_i$ over space, 2) its own tiny MLP $f_i$ generating both density and radiance values, 3) a single semantic logit vector $\bfs_i$ storing its semantic information directly, and 4) an associated instance ID $\text{Ins}_i$. Compared to other semantics-aware NeRF methods~\cite{Zhi21iccv_SemanticNeRF, vora2021nesf}, we use a single logit vector to represent local semantic information instead of training an MLP to encode semantics. This aligns with our goal that a single nerflet should not span multiple classes or instances, and has the additional benefits of reducing the capacity burden of the MLP and providing a natural inductive bias towards 3D spatial consistency.

\noindent\textbf{Pose and influence:}  Each nerflet has 9 pose parameters-- a 3D center $\bfmu_i$, 3 axis-aligned radii, and 3 rotation angles. We interpret these pose parameters in two ways. First, as a coordinate frame-- each nerflet can be rasterized directly for visualization by transforming an ellipsoid into the coordinate frame defined by the nerflet. This is useful for editing and understanding the scene structure (e.g., Figure~\ref{fig:teaser}).
The second way, more critical for rendering, is via an influence function $g_i(\bfx)$ defined by the same 9 pose parameters. $g_i$ is an analytic radial basis function (RBF) based on scaled anisotropic multivariate Gaussians~\cite{genova2019learning, genova2020local}
\begin{align}
\label{eq:influence_fun}
g_i(\bfx)=\eta\exp \left(-\frac{1}{2\tau}\left(\bfx-\bfmu_i\right)^T \Sigma_i^{-1}\left(\bfx-\bfmu_i\right)\right).
\end{align}
$\bfmu_i$ is the center of the basis function and $\Sigma_i$ is a 6-DOF covariance matrix. The covariance matrix is determined by 3 Euler-angle rotation angles and 3 axis-aligned radii that are the reciprocal of variance along each principal axis. These 9 parameters provide a fast and compact way to evaluate a region of influence for each nerflet without evaluating any neural network. This property is crucial for our fast training and evaluation, which will be introduced later.  $\eta$ is a scaling hyper-parameter set to 5 for all experiments, and $\tau$ is a scheduled temperature hyper-parameter used to control the falloff hardness. $\tau$ is reduced after each training epoch to minimize overlap between nerflets gradually.

\begin{figure}[t]
  \centering
  \includegraphics[width=0.85\linewidth]{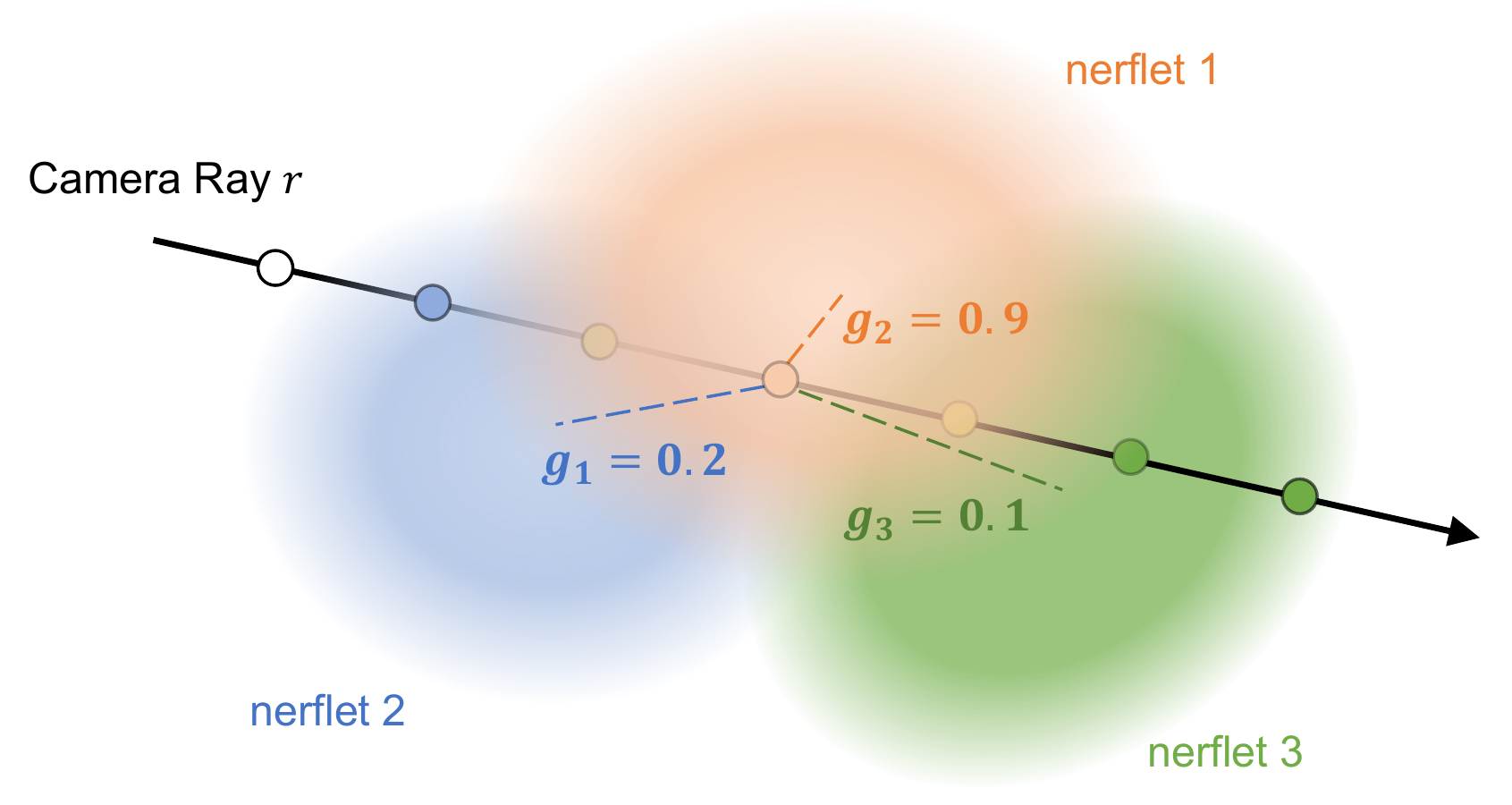}
  \vspace{-0.7em}
   \caption{Sample and blend method illustration. Results from individual nerflets are mixed based on the influence values $g_i$ determined by the distance-based weighting function. The mixing is smooth but most locations in space are dominated by a single nerflet, even when there is some overlap.}
   \label{fig:blend}
\end{figure}
\noindent\textbf{Rendering and blending:}
Given a scene represented by $N$ nerflets, we can render 2D images with volume rendering, as in NeRF:
\begin{align}
\label{eq:nerf_raymarching}
\hat{C}(\bfr) &=\sum_{k=1}^K T_k \alpha_k \bfc_k, \\
\text{where\ \ }
 \alpha_k &= 1-\exp \left(-\sigma_k \delta_k\right), \\
\label{eq:nerf_transmission}
 T_k &=\prod_{j=1}^{k-1} (1-\alpha_j).
\end{align}
$\hat{C}(\bfr)$ is the final color of ray $\bfr$, $T_k$ is transmission at the $k$-th sample along the ray, $\alpha_k$ is the opacity of the sample, $\bfc_k$ is the color at the sample, $\delta_k$ is the thickness of the current sample on the ray, and $\sigma_k$ is the density at the sample. We denote $k$ for the index of the sample along a ray, $K$ total number of samples and reserve $i$ for the index of the nerflet.

The biggest difference from NeRF is that instead of employing a single large MLP to produce $\bfc_k$ and $\sigma_k$, we combine values produced by individual nerflets using their influence weights (Figure~\ref{fig:blend}).
We query individual nerflet MLPs $f_i$ at the $k$-th input sample $(\text{pos}, \text{dir})=(\bfx_k, \bfd)$ along the ray, producing producing $N$ values $\sigma_{k,i}$ and $\bfc_{k,i}$ for nerflets labeled $i \in [1,N]$.
We then map the individual nerflet $\sigma_{k,i}$ values to $\alpha_{k,i}$ for rendering using $\delta_k$ and Equation~\ref{eq:nerf_raymarching}.
Finally, we take a weighted average of the $N$ individual nerflet color and $\alpha$ values to produce $\bfc_k$ and $\alpha_k$:
\begin{align}
    \label{eq:nerflet_render_color}
    \bfc_k &= \sum_{i=1}^{N}\hat{g}_i(\bfx_k)\bfc_{k,i}, \\
    \label{eq:nerflet_render_alpha}
    \alpha_k &= \sum_{i=1}^{N}\hat{g}_i(\bfx_k)\alpha_{k,i}, \\
    \label{eq:nerflet_render_g_norm}
    \text{where\ \ } \hat{g}_i(\bfx_k) &= \frac{g_i(\bfx_k)}{\sum_{j=1}^N{g_j(\bfx_k)+\epsilon}}.
\end{align}
$\epsilon$ is a factor allowing smooth decay to zero in empty space, with $\Sigma g_i(\bfx_k) \sim \epsilon$. 
After blending, the $\alpha_k$ and $\bfc_k$ values are used directly for ray marching as in Eq.~\ref{eq:nerf_raymarching}-\ref{eq:nerf_transmission} to generate final pixel color values $\hat{C}(\bfr)$, as in NeRF.

While in principle we should evaluate all nerflet MLPs in the scene in this step, as gaussian RBFs have infinite support, we do not do this. Typically, $g_i$ is dominated by one or at most a handful of nerflets that are close to the sample.
Therefore, we evaluate only the nearby MLPs, improving performance and scalability. This is implemented with our custom CUDA kernel, which will be introduced in Sec.~\ref{sec:effient_evaluation}. More distant nerflets are omitted from the average.

To generate semantic images, we average the per-nerflet semantic logit vector for each point sample in the same way described for color values $\bfc_{k,i}$ above.

To handle instances, we first compute a \emph{nerflet influence activation} function $\bfw \in \mathbb{R}^n$ for each point sample:
\begin{align}
    \label{eq:nerflet_influence_map}
    \bfw(\bfx) = \text{SoftMax}([\sigma_1g_1(\bfx),\dots,\sigma_Ng_N(\bfx)]).
\end{align}
$\sigma_i$ is the density evaluation for $\bfx$ on the $i$-th nerflet. This value intuitively represents how much influence each nerflet has over a given point sample, and can be accumulated by ray marching to generate a nerflet influence map $W(\bfr)$ for each ray $\bfr$. $W(\bfr)$ continuously captures which nerflet is dominant for each final pixel value, and is used by our influence loss described in Section~\ref{sec:loss_functions}. To assign a discrete instance label to a query position or ray, we take $\mathrm{argmax}_{i \in [N]} w(\bfr)$ or $\mathrm{argmax}_{i \in [N]} W(r)$, respectively to get $i$, then output $\text{Ins}_i$, the instance of that nerflet.


\noindent\textbf{Unbounded scenes:} Nerflets support both indoor (bounded) and outdoor (unbounded) scenes. To handle unbounded scenes, we add a single MLP $f_\text{far}$ to evaluate samples outside a large scene bounding box. We draw $M$ additional samples for these points at the end of the ray, which we concatenate after our blended RGB$\alpha$ values and composite. We use the scheme proposed by Zhang \etal~\cite{zhang2020nerf++}, with an added semantics branch, though as the content is very distant and nearly directional, many other approaches would likely work well (e.g., an environment map).

\subsection{Loss Function}
\label{sec:loss_functions}
During training, we jointly optimize all network parameters as well as the pose of the nerflets. In this way, each nerflet can be pushed by gradients to ``move'' across the scene, and focus on a specific portion of it.  We expect a final decomposition mirroring the scene, with more nerflets on complex objects, and use multiple losses to that end.

The loss function is broken up into rgb, semantic, instance, and regularization terms:
\begin{align}
    \label{eq:losses}
    \calL = \calL_\text{rgb} + \calL_\text{sem} + \calL_\text{ins} + \calL_\text{reg}. 
\end{align}

$\mathbf{\calL_{rgb}}$\textbf{:} The RGB loss $\calL_\text{rgb}$ is the mean squared error between the synthesized color $\hat{C}$ and the ground truth color $C$ averaged over a batch of sampled rays, as in the original NeRF. The one change is that we weight this loss with a schedule parameter that is 0.0 at step 0. We gradually increase this value to 1.0 during training to prevent early overfitting to high frequency appearance information.

$\mathbf{\calL_{sem}}$\textbf{:} Our semantic loss $\calL_\text{sem}$ compares the volume-rendered semantic logits pixel with the Panoptic Deeplab prediction~\cite{cheng2020panoptic}. We use a per-pixel softmax-cross-entropy function for this loss.

$\mathbf{\calL_{ins}}$\textbf{:} The instance loss is defined as:
\begin{align}
    \label{eq:instance_loss}
    \calL_\text{ins} = -\frac{1}{P}\sum_{(\bfr_1, \bfr_2)} ||W(\bfr_1) - W(\bfr_2)||_1.
\end{align}
That is, we sample $P$ ray pairs $(\bfr_1, \bfr_2)$ that are from the same class but different instances according to the instance segmentation model prediction, and enforce them to have different influence maps.
While this approach is somewhat indirect, well separated nerflets can be easily assigned instance labels (Section~\ref{sec:instance_assignment}), and it has the advantage of avoiding topology issues due to a variable number of instances in the scene while still achieving an analysis-by-synthesis loss targeting the instance decomposition. It is also compatible with the inconsistent instance ID labelings across different 2D panoptic image predictions.
Ray pairs $(\bfr_1,\bfr_2)$ are chosen within in an $L \times L$ pixel window per-batch for training efficiency.

$\mathbf{\calL_{reg}}$\textbf{:} 
Our regularization loss has several terms to make the structure of the nerflets better mirror the structure of the scene--
\begin{align}
    \label{eq:losses_reg}
    \calL_\text{reg} = \calL_\text{density} + \calL_\text{radii} + \calL_{\ell_1} + \calL_\text{box}. 
\end{align}
In addition to the intuitions described below, each of these is validated in a knock-out ablation study (Sec.~\ref{subsec:regularizers}) and tested on multiple datasets, to reduce the risk of overfitting to one setting.

First, to minimize unnecessary nerflet overlap within objects and reduce scene clutter, we penalize the $L_2$ norm of the radii of the nerflets ($\calL_{radii}$). To encourage sparsity where possible, we penalize the $L_1$ norm of the nerflet influence values at the sample locations ($\calL_{\ell_1}$). We also require nerflets to stay within their scene bounding box, penalizing:
\begin{align}
    \sum_{d \in \{x,y,z\}} \max(x_d - \mathrm{box}_{\mathrm{max}}, \mathrm{box}_{\mathrm{min}} - x_d, 0).
\end{align}
This reduces risk of ``dead'' nerflets, where a nerflet is far from the scene content, so it does not contribute to the loss, and therefore would receive no gradient.

Finally, we incorporate a ``density'' regularization loss $\calL_\text{density}$, which substantially improves the decomposition quality:
\begin{align}
    \label{eq:density_loss}
    \calL_\text{density} = -\frac{1}{D}\sum_{i=1}^n\sum\nolimits_{\bfx\sim\mathcal{N}(\bfmu_i, \Sigma_i)}\sigma(\bfx).
\end{align}
$\mathcal{N}(\bfmu_i, \Sigma_i)$ represents the underlying multivariate gaussian distribution for the $i$-th nerflet and $D$ is the number of samples drawn. This term rewards a nerflet for creating density near its center location. As a result, nerflets end up centered inside the objects they reconstruct.



\subsection{Instance Label Assignment}
\label{sec:instance_assignment}
Given an optimized scene representation, we use a greedy merge algorithm to group the nerflets and associate them with actual object instances. We first pick an arbitrary 2D instance image, and render the associated nerflet influence map $W(\bfr)$ for the image. We then assign the nerflets most responsible for rendering each 2D instance to a 3D instance based on it. We proceed to the next image, assigning nerflets to new or existing 3D instances as needed. Because the nerflets have been optimized to project to only a single 2D instance in the training images, this stage is not prone to failure unless the 2D panoptic images disagree strongly. See the supplemental for additional details.


\subsection{Efficient Nerflet Evaluation}
\label{sec:effient_evaluation}
\noindent\textbf{Top-$k$ Evaluation:} Instead of evaluating all nerflet MLPs in a scene as in Eq.~\ref{eq:nerflet_render_color}-\ref{eq:nerflet_influence_map}, we use the $g_i$ influence values to filter out distant and irrelevant nerflets in two ways. First, we truncate all nerflet influences below some trivial threshold to 0-- there is no need to evaluate the MLP at all if it has limited support. In free space, often all MLPs can be ignored this way. Next, we implement a ``top-$k$'' MLP evaluation CUDA kernel that is compatible with the autodiff training and inference framework. This kernel evaluates only the highest influence nerflet MLPs associated with each sample. We use $k=16$ for both training and visualizations in this paper, although even more aggressive pruning is quite similar in image quality (e.g., a difference of only $\sim0.05$ PSNR between top-16 and top-3) and provides a substantial reduction in compute. A top-$k$ ablation study is available in the supplemental.

\noindent\textbf{Interactive Visualization and Scene Editing:}
We develop an interactive visualizer combining CUDA and OpenGL for nerflets that takes advantage of their structure, efficiency, rendering quality, and panoptic decomposition of the scene. Details are available in the supplemental. The key insight enabling efficient evaluation is that nerflets have a good sparsity pattern for acceleration -- they are sparse with consistent local structure. We greatly reduce computation with the following two pass approach. In the first pass, we determine where in the volumetric sample grid nerflets have high enough influence to contribute to the final image. In the second pass, we evaluate small spatially adjacent subgrids for a particular nerflet MLP, which is generally high-influence for all samples in the subgrid due to the low spatial frequency of the RBF function. This amortizes the memory bandwidth of loading the MLP layers into shared memory. This approach is not as fast as InstantNGP~\cite{mueller2022instant}, but is still interactive and has the advantage of mirroring the scene structure.

\section{Experiments}
\label{sec:exp}

In this section, we evaluate our method using 512 nerflets on multiple tasks with two challenging real-world datasets. Please see the supplemental for other hyperparameters.

\noindent\textbf{KITTI-360:}
For KITTI-360 experiments, we use the novel view synthesis split, and compare to Panoptic Neural Fields (PNF)~\cite{kundu2022panoptic}, a recent state of the art method for panoptic novel view synthesis (1st on the KITTI-360 leaderboard).  To generate 2D panoptic predictions for outdoor scenes, we use a Panoptic DeepLab~\cite{cheng2020panoptic} model trained on COCO~\cite{lin2014microsoft}.

\noindent\textbf{ScanNet:} For ScanNet experiments, we evaluate on 8 scenes as in DM-NeRF~\cite{wang2022dmnerf} and compare to recent baselines-- DM-NeRF~\cite{wang2022dmnerf}, which synthesizes both semantics and instance information, and Semantic-NeRF~\cite{Zhi21iccv_SemanticNeRF} which synthesizes semantics only.  To generate panoptic images for indoor scenes, we use PSPNet~\cite{zhao2017pyramid} and Mask R-CNN~\cite{he2017mask}. Please see the supplemental for important subtleties regarding how 2D supervision on ScanNet is achieved.

\begin{table}
\centering
\resizebox{\linewidth}{!}{ 
\begin{tabular}{lccc}
\hline
\multicolumn{1}{p{2cm}}{Method}                                                           & \multicolumn{1}{p{2cm}}{\centering Appearance \\ PSNR}  & \multicolumn{1}{p{2cm}}{\centering Semantics \\ mIOU} & \multicolumn{1}{p{2cm}}{\centering Worst Case \\ kFLOPs} \\ \hline
PBNR~\cite{kopanas2021point} + PSPNet~\cite{zhao2017pyramid}     & 19.91  & 65.07  &  - \\
FVS~\cite{Riegler2020ECCV} +  PSPNet~\cite{zhao2017pyramid}      & 20.00  & 67.08  &  - \\
NeRF~\cite{mildenhall2021nerf} + PSPNet~\cite{zhao2017pyramid}   & 21.18  & 53.01  &  $\sim$ 1056\\
Mip-NeRF~\cite{barron2021mip} + PSPNet~\cite{zhao2017pyramid}    & 21.54  & 51.15  &  $\sim$ 1056 \\
PNF~\cite{kundu2022panoptic}                  & \textbf{21.91}  & \underline{74.28} & $\sim$ 1256  \\
Ours                                                             & \underline{21.69}  & \textbf{75.07} & $\sim$ 244 \\ \hline
\end{tabular}
} 
   \vspace{-0.8em}
\caption{
Results on novel view color and semantic synthesis tasks on KITTI-360~\cite{kitti360}. Nerflets achieve similar color synthesis quality and better semantic synthesis quality compared to PNF~\cite{kundu2022panoptic} without any 3D supervision. We also have the best efficiency in terms of worst case kFLOPs.
} 
\label{tab:kitti360}
\end{table}
\begin{figure}[t]
  \centering
  \includegraphics[width=\linewidth]{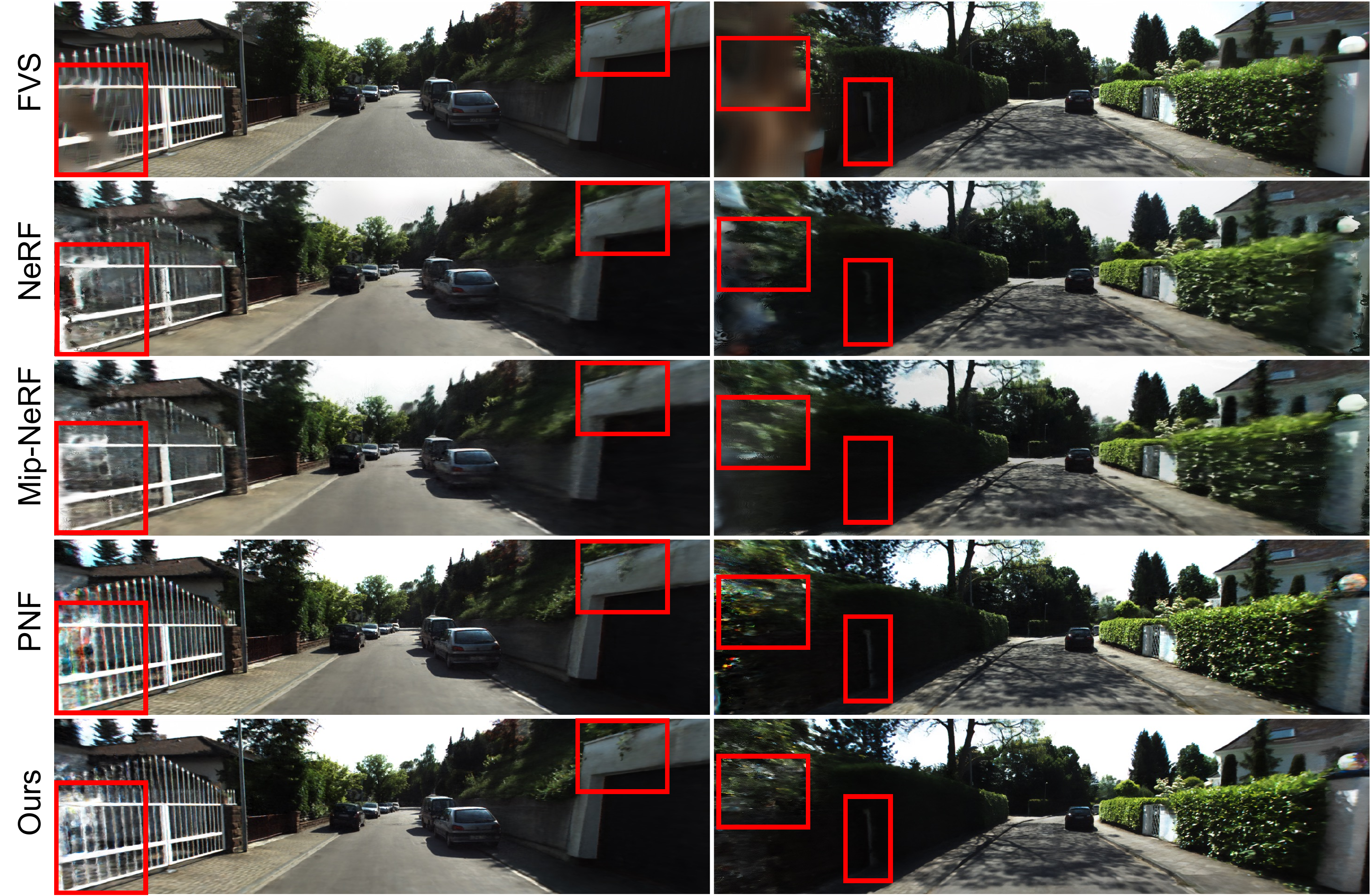}
   \vspace{-1.8em}
   \caption{Novel view synthesis qualitative comparison. Nerflets outperform NeRF, Mip-NeRF, and FVS, and perform comparably to PNF with better performance in difficult areas (far left), possibly due to explicit spatial allocation of parameters.}
   \label{fig:kitti360_photo}
\end{figure}
\begin{figure}[t]
  \centering
  \includegraphics[width=\linewidth]{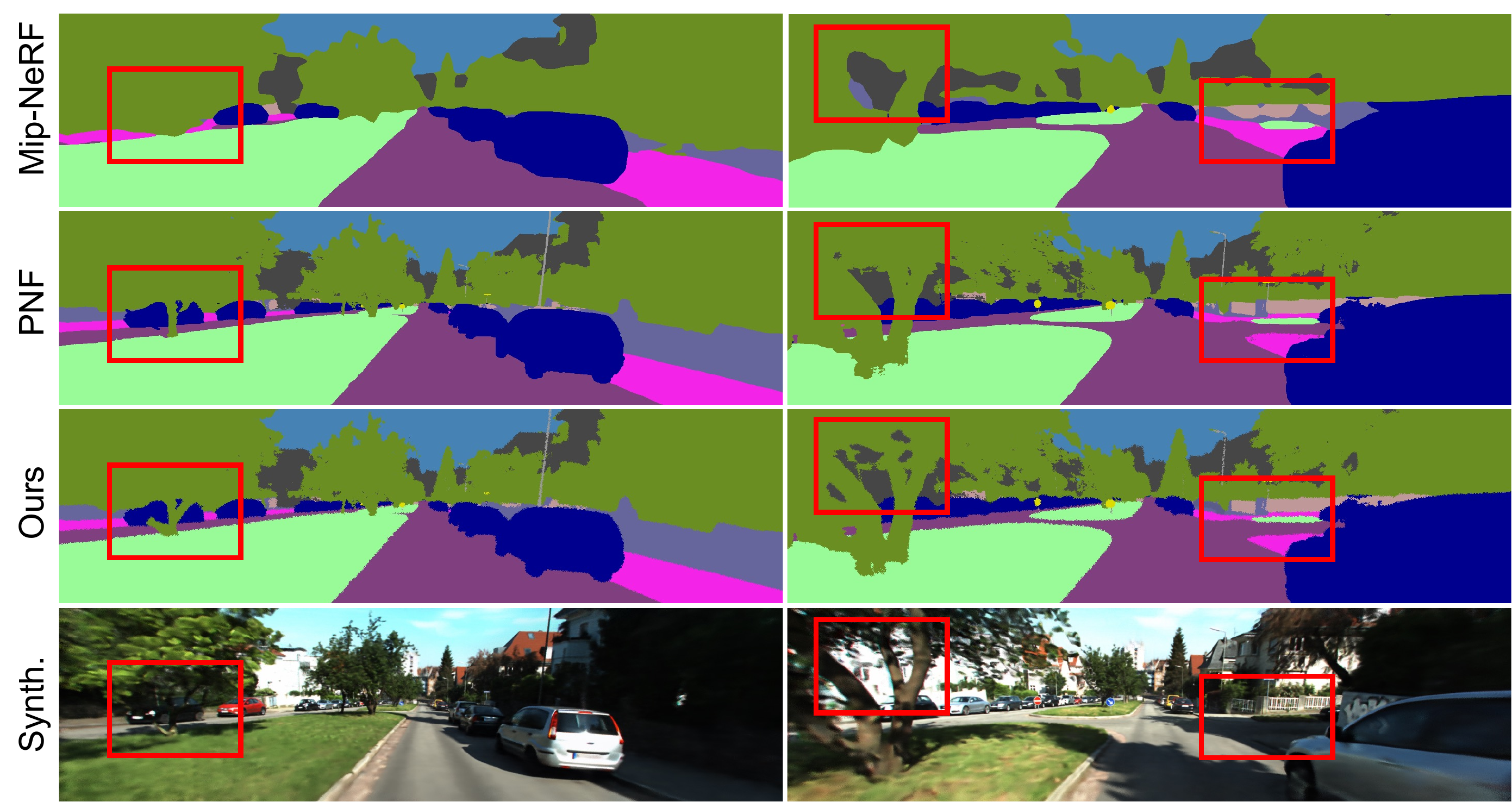}
     \vspace{-1.8em}
   \caption{Novel view semantic synthesis qualitative comparison. Nerflets outperform other methods, particularly with respect to details and thin structures.}
   \label{fig:kitti360_semantic}
\end{figure}
\begin{figure}[t]
  \centering
  \includegraphics[width=\linewidth]{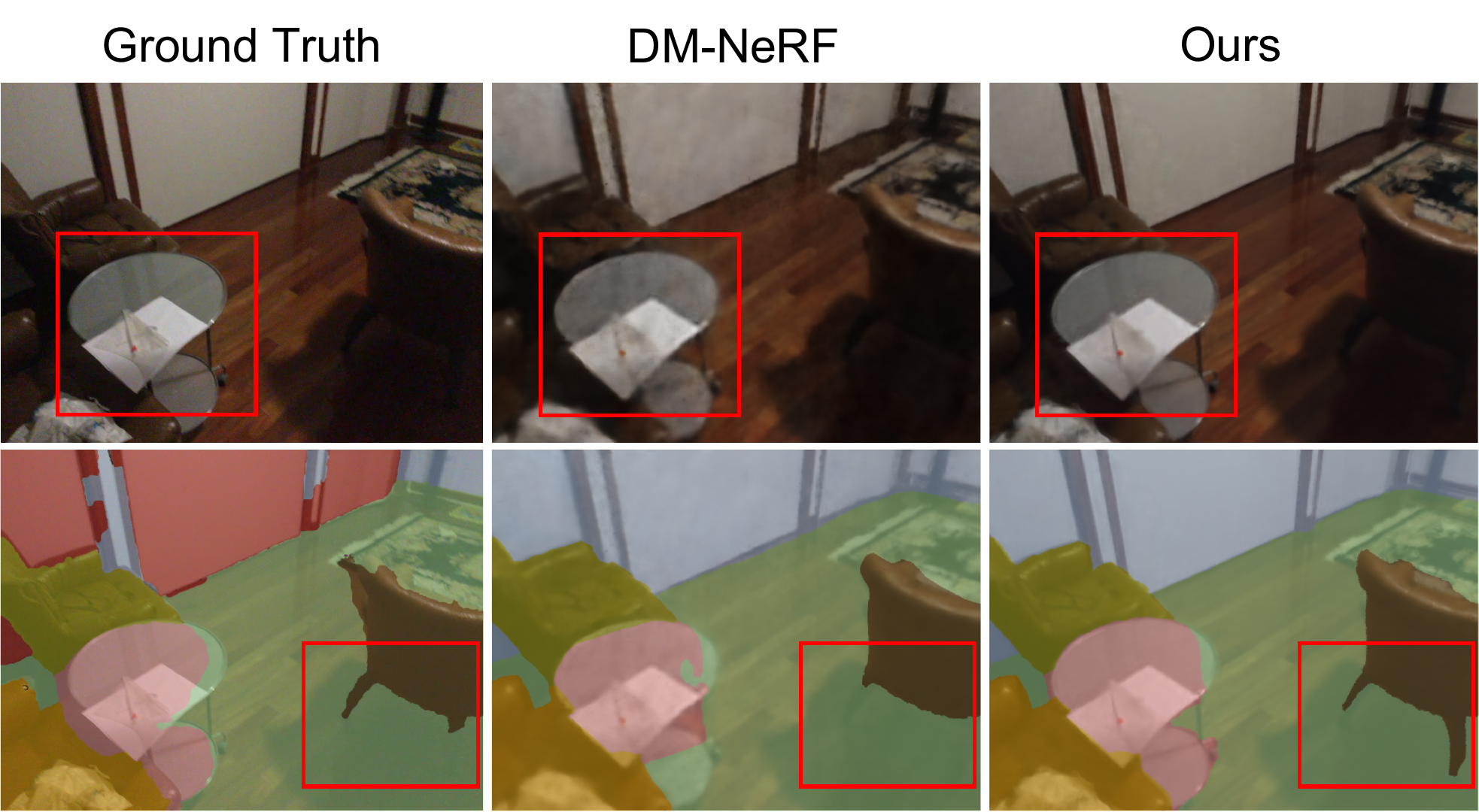}
  \vspace{-1.8em}
   \caption{ScanNet qualitative result and comparison to DM-NeRF~\cite{wang2022dmnerf}. The comparison is on a ScanNet view synthesis example taken from the DM-NeRF paper.  Our results improve in terms of both image and segmentation quality- notice the better image rendering for the glass table, and the better segmentation of the chair legs, which even exceeds the ground truth quality.}
   \label{fig:scannet_nvs}
\end{figure}

\begin{figure}[t]
  \centering
  \includegraphics[width=1.0\linewidth]{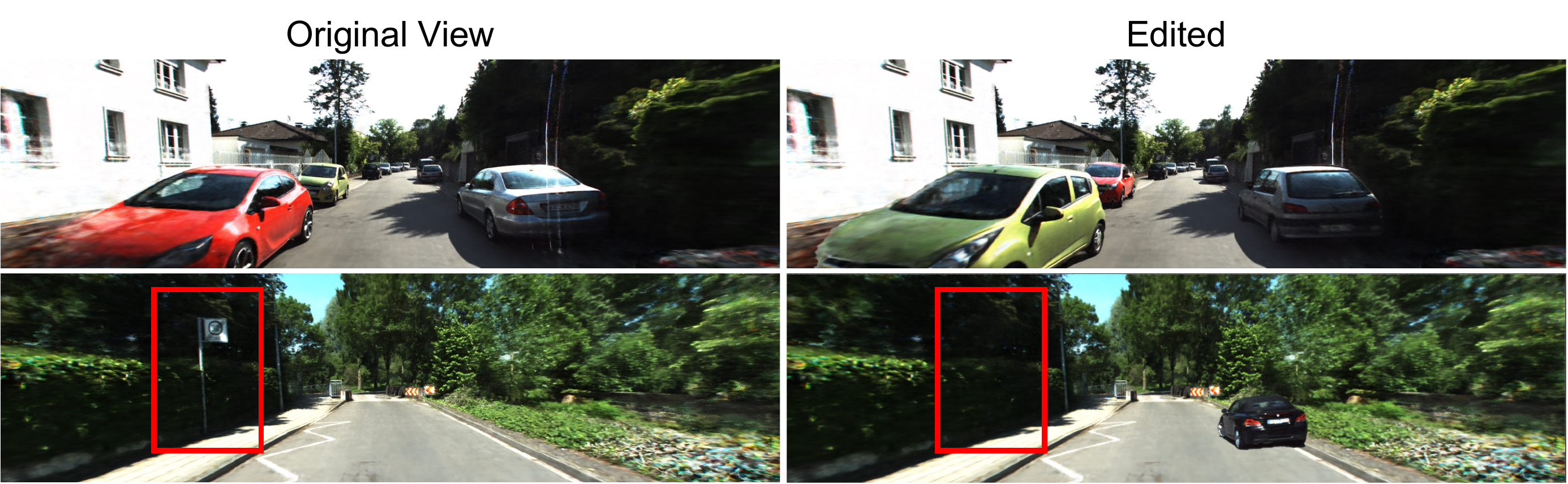}
  \vspace{-1.8em}
   \caption{KITTI-360 scene editing. We replace cars (top) or removing a sign instance (bottom).}
   \label{fig:kitti360_editing}
\end{figure}\textbf{}

\begin{table}
\centering
\vspace{-0.3em}
\resizebox{\linewidth}{!}{ 
\begin{tabular}{lccc}
\hline
Method                                            & PSNR  & Semantics mIoU  & Instance mAP\footnotesize{0.5}\\ \hline
PSPNet~\cite{zhao2017pyramid} on GT Image                     & -     & 68.43  & -     \\
Mask R-CNN~\cite{he2017mask}  on GT Image                      & -     & -     & 23.53  \\
\hline
PSPNet~\cite{zhao2017pyramid} on NeRF Im.                     & -     & 46.21  & -     \\
Mask R-CNN~\cite{he2017mask}  on NeRF Im.                      & -     & -     & 14.32  \\
Semantic-NeRF~\cite{Zhi21iccv_SemanticNeRF}       & 28.43  & 71.34  & -     \\
DM-NeRF~\cite{wang2022dmnerf}                     & 28.21  & 70.71  & 25.12  \\
Ours                                              & \textbf{29.12}  & \textbf{73.63}  & \textbf{31.32}  \\  \hline
\end{tabular}
} 
\vspace{-0.8em}
\caption{
ScanNet novel view synthesis quantitative results. Rendered color images and segmentation maps from nerflets have the best quality among all evaluated methods.
} 
\label{tab:scannet}
\end{table}
\begin{figure}[t]
  \centering
    \vspace{-0.4em}
  \includegraphics[width=1.0\linewidth]{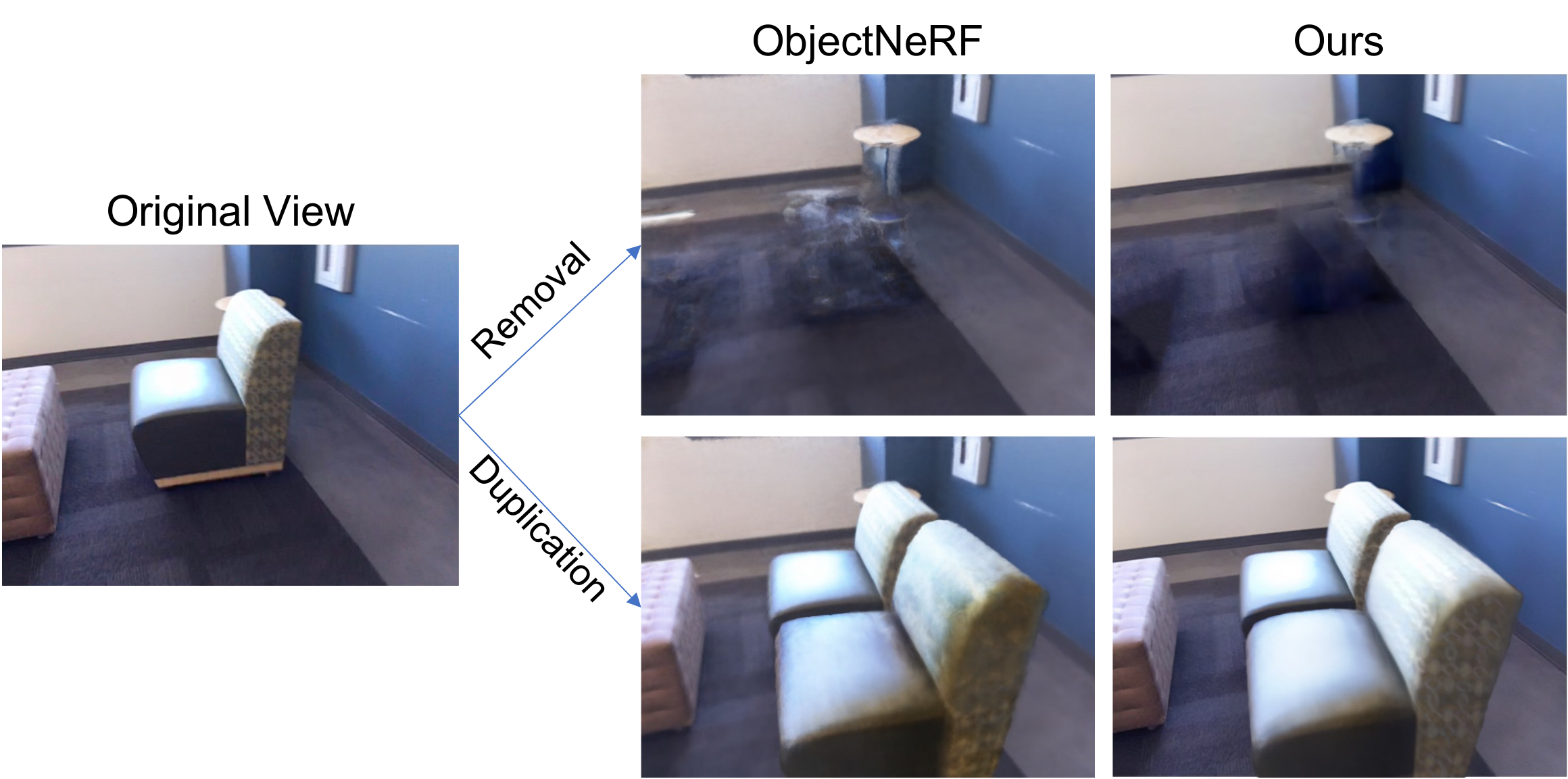}
  \vspace{-1.8em}
   \caption{Scene editing comparison with ObjectNeRF on a pair of ScanNet images shown in the ObjectNeRF paper. Notice the improved handling of free space during removal and the more accurate texture during duplication, both attributable to the simple ``copy-and-paste'' nature of nerflets manipulation. }
   \label{fig:scannet_editing}
\end{figure}
\begin{figure}[t]
  \centering
  \vspace{-0.3em}
  \includegraphics[width=\linewidth]{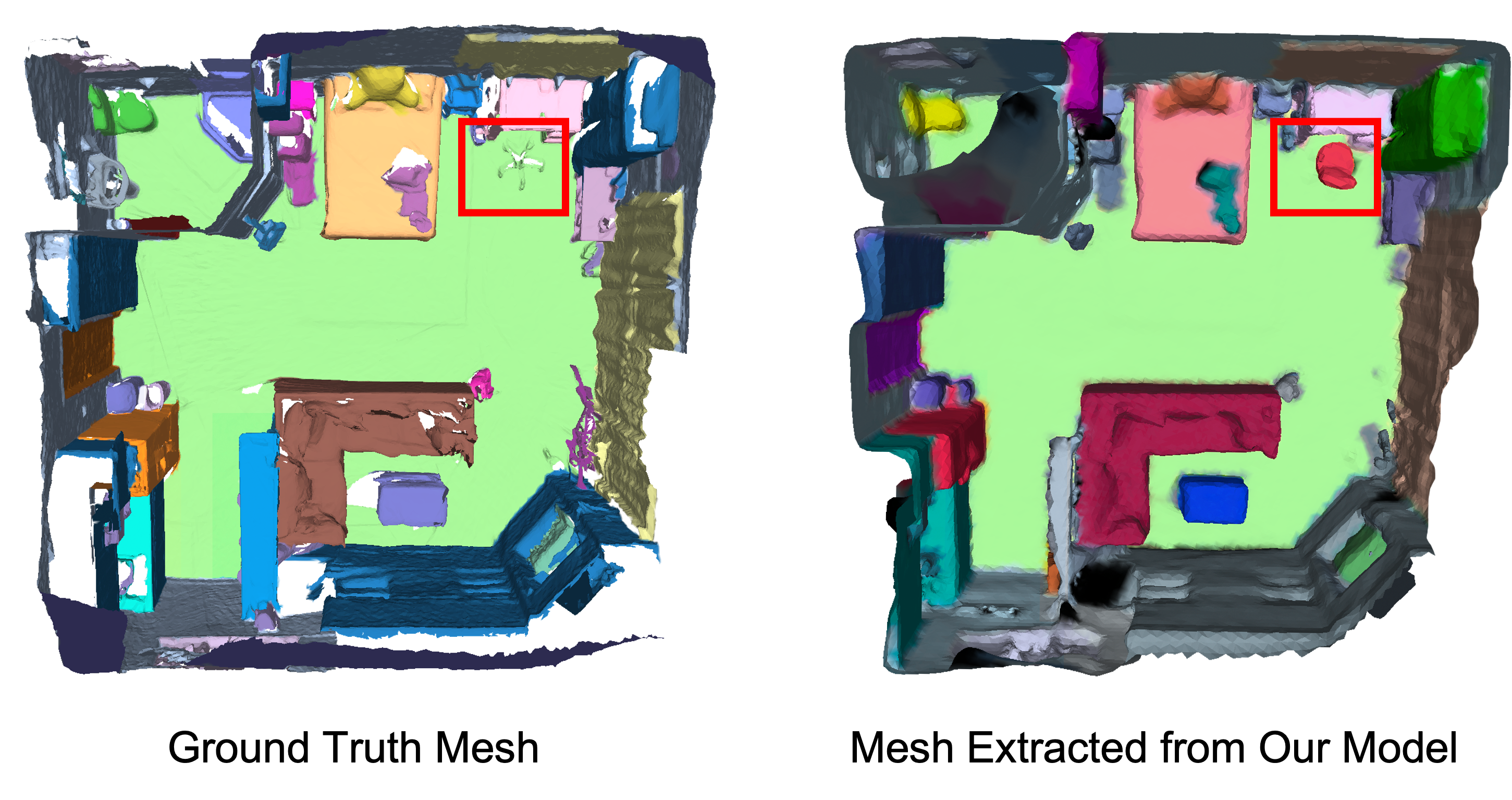}
  \vspace{-2.0em}
   \caption{A 3D mesh extracted from a ScanNet RGB sequence with nerflets, colorized according to predicted panoptic labels. Ground truth RGBD mesh with human-annotated labels is shown on the left. Nerflets successfully reconstruct and label a chair instance missing from the ground truth mesh.}
   \label{fig:scannet_3d}
\end{figure}

\begin{table}
\centering
\resizebox{0.95\linewidth}{!}{ 
\begin{tabular}{c|ccc}
\hline
Supervision          & Method           & Semantics mIoU & Instance mAP\footnotesize{0.5} \\ \hline
\multirow{5}{*}{Pointcloud}           & MinkowskiNet~\cite{choy20194d}               & \textbf{71.92}          & -            \\
                              & 3DMV~\cite{dai20183dmv}                      & 49.22          & -            \\
                              & PointNet++~\cite{qi2017pointnet++}           & 44.54          & -            \\
                              & Mask3D~\cite{Schult22}                       & -              & \textbf{75.34}        \\
                              & 3D-BoNet~\cite{yang2019learning}             & -              & 46.23        \\ \hline
\multirow{2}{*}{Images}           & Multiview Fusion~\cite{genova2021learning}   & 55.23           & -            \\
                              & Ours                                         & \textbf{63.94}          & 48.67        \\ \hline
\end{tabular}
}
\vspace{-0.8em}
\caption{
Evaluation of nerflet panoptic performance on ScanNet 3D point cloud labeling task. Nerflets beat some less recent fully 3D-supervised methods with less supervision at both semantic and instance tasks, while also beating the similarly supervised multiview fusion approach.
}
\label{tab:scannet3d}
\end{table}
\begin{figure*}[t]
  \centering
  \vspace{-0.5em}
  \includegraphics[width=\linewidth]{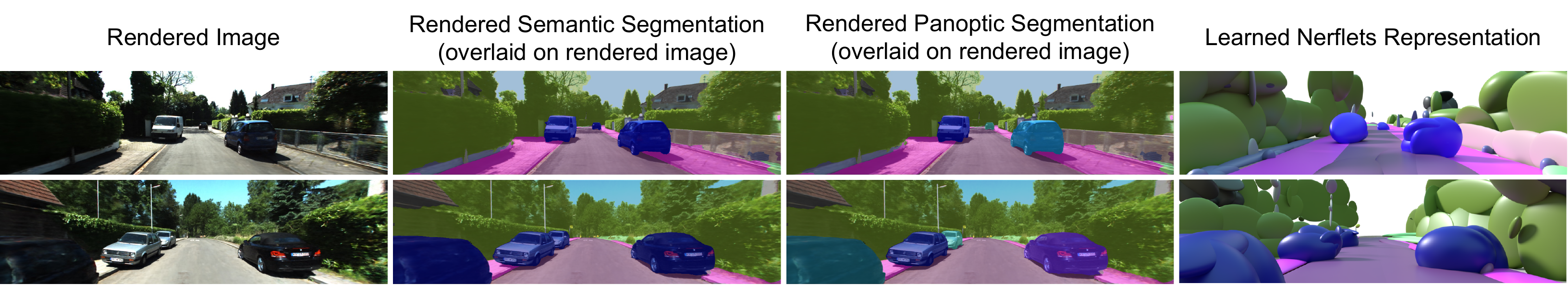}
  \vspace{-2.0em}
   \caption{Visualization of nerflet outputs, trained on KITTI-360 images.}
   \label{fig:kitti360_panoptic}
\end{figure*}

\vspace{-1em}
\subsection{Results}
\noindent\textbf{Novel View Synthesis:} 
We evaluate the performance of nerflets for novel view image synthesis on both KITTI-360 and ScanNet. As shown in Table~\ref{tab:kitti360}, on KITTI-360, our method achieves better PSNR than all other 2D supervised methods on the leaderboard and is competitive with PNF~\cite{kundu2022panoptic}, which utilizes 3D supervision. As shown in Figure~\ref{fig:kitti360_photo} our visual quality is approximately on par with PNF, and does particularly well in challenging areas. For complex indoor scenes in ScanNet (Table~\ref{tab:scannet}), we achieve the best performance for novel view synthesis under all settings (with or without instance supervision), including when compared to DM-NeRF~\cite{wang2022dmnerf}. In particular, nerflets achieve better object details (Figure~\ref{fig:scannet_nvs}), likely due to their explicit allocation of parameters to individual object instances. 

\noindent\textbf{2D Panoptic Segmentation:}
\label{sec:exp_panop2d}
Nerflets can render semantic and instance segmentations at novel views. We evaluate our 2D panoptic rendering performance quantitatively on both KITTI-360 (Table~\ref{tab:kitti360}) and ScanNet (Table~\ref{tab:scannet}). On both datasets, nerflets outperform all baselines in terms of semantic mIoU, even compared to the 3D-supervised PNF~\cite{kundu2022panoptic}. On the ScanNet dataset, we also show that nerflets outperform PSPNet~\cite{zhao2017pyramid} and Mask R-CNN~\cite{he2017mask} in terms of both mIoU and instance mAP, although those methods were used to generate the 2D supervision for our method. This is an indication that nerflets are not only expressive enough to represent the input masks despite their much lower-dimensional semantic parameterization, but also that nerflets are effectively fusing 2D information from multiple views into a better more consistent 3D whole. We further explore this in the supplemental material. Qualitatively, nerflets achieve better, more detailed segmentations compared to baseline methods (Figure~\ref{fig:kitti360_semantic}, Figure~\ref{fig:scannet_nvs}), particularly for thin structures.

\noindent\textbf{Scene Editing:}
\label{subsec:exp_editing}
In Figure~\ref{fig:kitti360_editing} and Figure~\ref{fig:scannet_editing}, we use the instance labels on nerflets to select individual objects, and then manipulate the nerflet structure directly to edit scenes. 
No additional optimization is required, and editing can be done while rendering at interactive framerates (please see the video for a demonstration, the results here were rendered by the standard autodiff inference code for the paper).
When compared to Object-NeRF on ScanNet (Figure~\ref{fig:scannet_editing}), nerflets generate cleaner results with more detail, thanks to their explicit structure and alignment with object boundaries. Using nerflets, empty scene regions will not carry any density after deletion, as there is nothing there to evaluate. In Figure~\ref{fig:kitti360_editing}, we demonstrate additional edits on KITTI-360, with similar results. These visualizations help to confirm that indeed, nerflets learn a precise and useful 3D decomposition of the scene.

\noindent\textbf{3D Panoptic Reconstruction:}
In Figure~\ref{fig:scannet_3d}, we demonstrate the 3D capabilities of nerflets by extracting a panoptically labeled 3D mesh and comparing it to the ground truth. Surface extraction details are provided in the supplemental. We observe that the resulting mesh has both good reconstruction and panoptic quality compared to the ground truth. For example, nerflets even reveal a chair instance that is entirely absent in the ground truth mesh. We demonstrate this quantitatively by transferring nerflet representations to a set of ground truth 3D ScanNet meshes, comparing to existing 3D-labeling approaches in Table~\ref{tab:scannet3d}. We observe that nerflets outperform the similarly supervised multi-view fusion baseline, while adding instance capabilities. State of the art directly-3D supervised baselines are still more effective than nerflets when input geometry and a large 3D training corpus are available, but even so nerflets outperform some older 3D semantic and instance segmentation methods.


\vspace{-0.5em}

\subsection{Analysis \& Ablations}
\label{sec:discussion}

\noindent\textbf{Scene Decomposition Quality:}
Our insight was to create an irregular representation that mirrors the structure of the scene. Do nerflets succeed at achieving this scene decomposition? In Fig.~\ref{fig:kitti360_panoptic}, we show RGB and panoptic images alongside the underlying nerflet decomposition that generated them. We see that indeed, the nerflets do not cross object boundaries, do join together to represent large or complex objects, and do cover the scene content.




\noindent\textbf{Semantics Help Appearance:}
\label{subsec:semvsapp}
One key insight about our approach is that the semantic structure of the nerflets decomposition is beneficial even for lower level tasks, like novel view synthesis. We perform an experiment on the KITTI-360 validation set and observe that when training without a semantic or instance loss (i.e., photometric and regularization losses only), nerflets achieve a PSNR of 20.95. But when adding the semantics loss, PSNR \textit{increases} to 22.43, because the nerflets end up more accurately positioned where the content of the scene is. This is also why we train with a higher semantic loss early in training, to encourage better nerflet positioning.

\begin{table}
\centering
\vspace{-0.6em}
\begin{tabular}{lccc}
\hline
                               & PSNR             & mIOU              &mAP\footnotesize{0.5}\\ \hline
w/o $\calL_\text{density}$     & 20.85            & 63.31             & 11.20  \\
w/o $\calL_\text{radii}$       & 27.23            & 72.43             & 26.32  \\ 
w/o $\calL_{\ell_1}$           & 28.83            & 68.23             & 21.74  \\
w/o $\calL_\mathrm{box}$       & 28.93            & 72.14             & 29.88  \\  
full model                     & \textbf{29.12}   & \textbf{73.63}    & \textbf{31.32}  \\

\hline
\end{tabular}
\vspace{-0.8em}
\caption{
Ablation experiment on ScanNet for the effectiveness of our regularization terms-- density loss $\calL_\text{density}$, radii penalty $\calL_\text{radii}$, influence sparsity loss $\calL_{\ell_1}$ and scene box loss. $\calL_\mathrm{box}$
} 
\label{tab:abl_reg}
\end{table}
\noindent\textbf{Ablation Study:}
\label{subsec:regularizers} 
Here we run a knock-out ablation study to validate the effectiveness of each of our regularization terms. Table~\ref{tab:abl_reg} shows that all regularization terms contribute to final performance quantitatively. $\calL_\text{density}$ is the most crucial term for learning a nice representation. It affects both image synthesis and segmentation performance, as it encourages nerflets to focus around actual scene content. $\calL_\text{radii}$, $\calL_{\ell_1}$ and $\calL_\mathrm{box}$ all also improve performance, due to their effect of forcing a more well-separated and active decomposition of the scene where all nerflets contribute to the final result.

\noindent\textbf{Performance:}
Nerflets have good performance due to their local structure. Our editor renders $320\times240$ top-$1$ editable volume images with 192 samples/pixel at 31 FPS with 4 A100 GPUs and 64 nerflets-- 457 million sample evaluations per second. 

\vspace{-0.5em}


\section{Conclusion and Limitations}
\label{sec:conclusion}
\vspace{-0.5em}
In this work, we present nerflets, a novel 3D scene representation which decomposes the scene into a set of local neural fields. Past work demonstrated structure is useful for parsimony in MLP-based shape representation~\cite{genova2020local}, and we have found similar evidence extending that to scenes in this paper.
Thanks to the locality of each nerflet, our model is compact, efficient, and multi-view-consistent. Results of experiments on two challenging real-world datasets KITTI-360 and ScanNet demonstrate state-of-the-art performance for panoptic novel view synthesis, as well as competitive novel view synthesis and support for downstream tasks such as scene editing and 3D segmentation. 

Despite these positives, nerflets have several limitations. For example, we do not model dynamic content. Even though the representation is well-suited for handling rigid motions (as demonstrated in scene editing), that feature has not been investigated. Also, while individual nerflet radiance fields are capable of handling participating media, the overall representation may struggle to fit scenes where those effects cross semantic boundaries (e.g., foggy outdoor sequences). Finally, we currently assume a fixed number of nerflets for each scene, regardless of the scene complexity. However, it may be advantageous to prune, add, or otherwise dynamically adjust the number based on where they are needed (\eg where the loss is highest).  
Investigating these novel features is interesting for future work.

\pagebreak
\noindent\textbf{Acknowledgements.} We sincerely thank Yueyu Hu, Nilesh Kulkarni, Songyou Peng, Mikaela Angelina Uy, Boxin Wang, and Guandao Yang for useful discussion and help on experiments. We also thank Avneesh Sud for feedback on the manuscript.

{\begin{flushleft}\LARGE \textbf{Appendix} \end{flushleft}}
\setcounter{section}{0}
\renewcommand\thesection{\Alph{section}}
\section{More Details}
\subsection{Model Architecture}

For all nerflets MLPs $f_i$, we follow the NeRF architecture~\cite{mildenhall2021nerf} but reduce the number of hidden layers from 8 to 4, and reduce the number of hidden dimensions from 256 to 32. We also removed the shortcut connection in the original network. All other architecture details are as in~\cite{mildenhall2021nerf}. The background neural field uses NeRF++~\cite{zhang2020nerf++} style encoding, and its MLP $f_\text{far}$ is with 6 hidden layers and 128 hidden dimensions. One distinction is that we do perform coarse-to-fine sampling as in~\cite{mildenhall2021nerf}, but both coarse and fine samples are drawn from a single MLP, not two distinct ones.

\subsection{Hyper-parameters}

We use $N=512$ nerflets for all experiments in the main paper. The scaling parameter $\eta$ is set to $5$ for all experiments. We initialize the nerflets temperature parameter $\tau$ to $1$ and multiply $\tau$ by $0.9$ across the epochs. The smooth decay factor $\epsilon$ is set to $10^{-7}$ for all experiments. We draw $64$ samples for coarse level and $128$ samples for fine level within the bounding box. For unbounded scenes, we draw $16$ coarse samples and $16$ fine samples from the background MLP. We increase the weight for $\calL_\text{rgb}$ from $0.0$ to a maximum of $1.0$ by the step of $0.2$ across epochs to prevent early overfitting to high frequency information. Contrastive ray pairs are sampled within an $32\times32$ pixel window. The weight for regularization loss $\calL_\text{reg}$ is set to $0.1$. All other losses are with weight 1.0.

\subsection{Dataset Details}
For training on each ScanNet scene, we uniformly sample $20\%$ of the RGB frames for training and $10\%$ of the RGB frames for evaluation-- about $200$ frames for training and $100$ frames for evaluation. For both ScanNet and KITTI-360 scenes, we estimate the scene bounding box with camera extrinsics and normalize the coordinate inputs to $[-0.5, 0.5]$ for all experiments.

One important note about the ScanNet~\cite{dai2017scannet} experiments is that 2D ScanNet supervision indirectly comes from 3D. That is because the 2D ScanNet dataset was made by rendering the labeled mesh into images. We do not use this 2D ground truth directly, but PSPNet~\cite{zhao2017pyramid} is trained on it. Here, this is primarily a limitation of the evaluation rather than the method-- there are many 2D models that can predict reasonable semantics and instances on ScanNet images, but we want to be able to evaluate against the exact classes present in the 3D ground truth. This does not affect the comparison to other 2D supervised methods, as all receive their supervision from the same 2D model. By comparison KITTI-360 results are purely 2D only, but all quantitative evaluations must be done in image-space.

\subsection{Paper Visualization Details}
For ScanNet mesh extraction, we create point samples on a grid and evaluate their density, semantic and instance information from nerflets. We then estimate point normals using the $5$ nearest neighbors and create a mesh with screened Poisson surface reconstruction~\cite{kazhdan2013screened}. The mesh triangles are colored according to the semantic and instance labels of their vertices. For the teaser and KITTI-360 visualizations of our learned nerflets representation, we visualize nerflets according to its influence function. We draw ellipsoids at influence value $e^{-\frac{1}{2}} \approx 0.607$.

\subsection{Interactive Visualizer Details}
Our interactive visualizer allows real-time previewing of nerflet editing results while adjusting the bounding boxes of objects in the scene. The visualizer draws the following components. First, a volume-rendered RGB or depth image at an interactive resolution of up to 320x240. This enables viewing the changes being made to the scene in real time. Second, the nerflets directly, by rendering an ellipsoid per nerflet at a configurable influence threshold. This enables seeing the scene decomposition produced by the nerflets. Third, a dynamic isosurface mesh extracted via marching cubes that updates as the scene is edited, giving some sense of where the nerflets are in relation to the content of the scene. Fourth, a set of boxing box manipulators, one per object instance, with draggable translation and rotation handles. These boxes are instantiated by taking the bounding box of the ellipsoid outline meshes for all nerflets associated with a single instance ID. A transformation matrix that varies per instance is stored and pushed to the nerflets on each edit.

Most of the editor is implemented in OpenGL, with the volume rendering implemented as a sequence of CUDA kernels that execute asynchronously and are transferred to the preview window when ready. In the main paper we report performance numbers for top-1 evaluation, which is often the right compromise for maximizing perceived quality in a given budget (e.g., pixel count can be more important), though interactive framerates with top-16 or top-3 evaluation are possible at somewhat lower resolutions.

\subsection{Instance Label Assignment}
To assign instance labels for each nerflet, we render the nerflet influence map $W_i$ for each view and compare with corresponding 2D semantic and instance segmentation maps to match each $l_i^j$ (2D object instance or stuff with local id $j$ in view $i$) to a set of nerflets $M(l_i^j)$. Here $M(\cdot)$ maps an instance ID to its set of associated nerflets. We then create a set of 3D instances $G=\{g_k\}$ according to the segmentation result of the first view -- we create a 3D global instance for each detected 2D object instance and also each disjoint stuff labels in the semantic maps. For each new view $i$, we match $l_i^j$ to the 3D instance $g_k$ if $|M(l_i^j) \cap M(g_k)| / |M(l_i^j)| \geq \delta$, and then update $M(g_k)$ to $M(g_k) \cup M(l_i^j)$. If no match is found in $\{g_k\}$, we create a new 3D instance and insert it into $G$. Before inserting any new global instance $g$, we remove all nerflets that are already covered by the global set $G$ from $g$. By using this first-come-first-serve greedy strategy we always guarantee no nerflet is associated with 2 different global instances. After this step, each nerflet is associated with a global instance ID, and our representation can be used to reason at an instance level effectively.


\section{More Results}
The following experiments are done on the subset of ScanNet from~\cite{wang2022dmnerf}.


\subsection{Number of Nerflets}
\begin{table}
\centering
\vspace{-0.6em}
\begin{tabular}{lccc}
\hline
          & PSNR         & mIOU    \\ \hline
$n=64$        & 26.34        & 53.23   \\
$n=128$       & 28.34        & 62.41   \\  
$n=256$       & 28.81        & 69.97   \\
$n=512$       & 29.12        & 73.63   \\
$n=1024$      & 29.19        & 74.09   \\

\hline
\end{tabular}
\vspace{-0.8em}
\caption{Ablation experiment on ScanNet for different number of nerflets.} 
\label{tab:supp_n_nerflets}
\end{table}

We perform ablation study on the number of nerflets on ScanNet. The results are in Tab.~\ref{tab:supp_n_nerflets}. We find that increasing the number of nerflets could improve the performance on both photometric and semantic metrics. However, the benefit saturates when adding more nerflets than 512. To balance performance and efficiency, we use 512 nerflets in all experiments in the paper.

\subsection{Effect of Top-$k$ Evaluation}
\begin{table}
\centering
\vspace{-0.6em}
\begin{tabular}{lccc}
\hline
          & PSNR         & mIOU      \\ \hline
$k=32$    & 29.13        & 73.72     \\  
$k=16$    & 29.12        & 73.63     \\
$k=3$     & 29.05        & 72.95     \\
$k=1$     & 28.35        & 70.73     \\

\hline
\end{tabular}
\vspace{-0.8em}
\caption{Ablation experiment on ScanNet for evaluating only nerflets with top-$k$ influence weights during training and testing.} 
\label{tab:supp_topk}
\end{table}

We perform ablation study on the performance impact of $k$ when we only evaluate nerflets with top-$k$ influence weights for each point sample. The results are in Tab.~\ref{tab:supp_topk}. We find that evaluating 32, 16 or 3 nerflets have little influence on the model performance, since each nerflet is only contributing locally. However we see a moderate performance drop when only evaluating one nerflets with the highest influence weight. We choose to evaluate $k=16$ nerflets for all experiments in the paper to balance the computational cost and performance.

\subsection{Inactive Nerflets}
One known problem~\cite{genova2019learning} with training using RBFs that have learned extent is that when an RBF gets too small or too far from the scene, it does not contribute to the construction results. The radii loss $\calL_\text{radii}$ and box loss $\calL_\text{box}$ are proposed to alleviate this issue. To estimate the actual number of inactive nerflets, we utilize nerflet influence map $W$ and count nerflets that do not appear on any of these maps in any views. In KITTI-360 experiments, we estimate to have 10.6 inactive nerflets on average per scene, making up $\sim 2.07\%$ of all available nerflets. In ScanNet experiments in the main paper, we estimate to have 30.6 inactive nerflets on average per scene, making up $\sim 5.98\%$ of all available nerflets.

\subsection{Robustness against Input 2D Segmentation}
\begin{figure}[t]
  \centering
  \includegraphics[width=\linewidth]{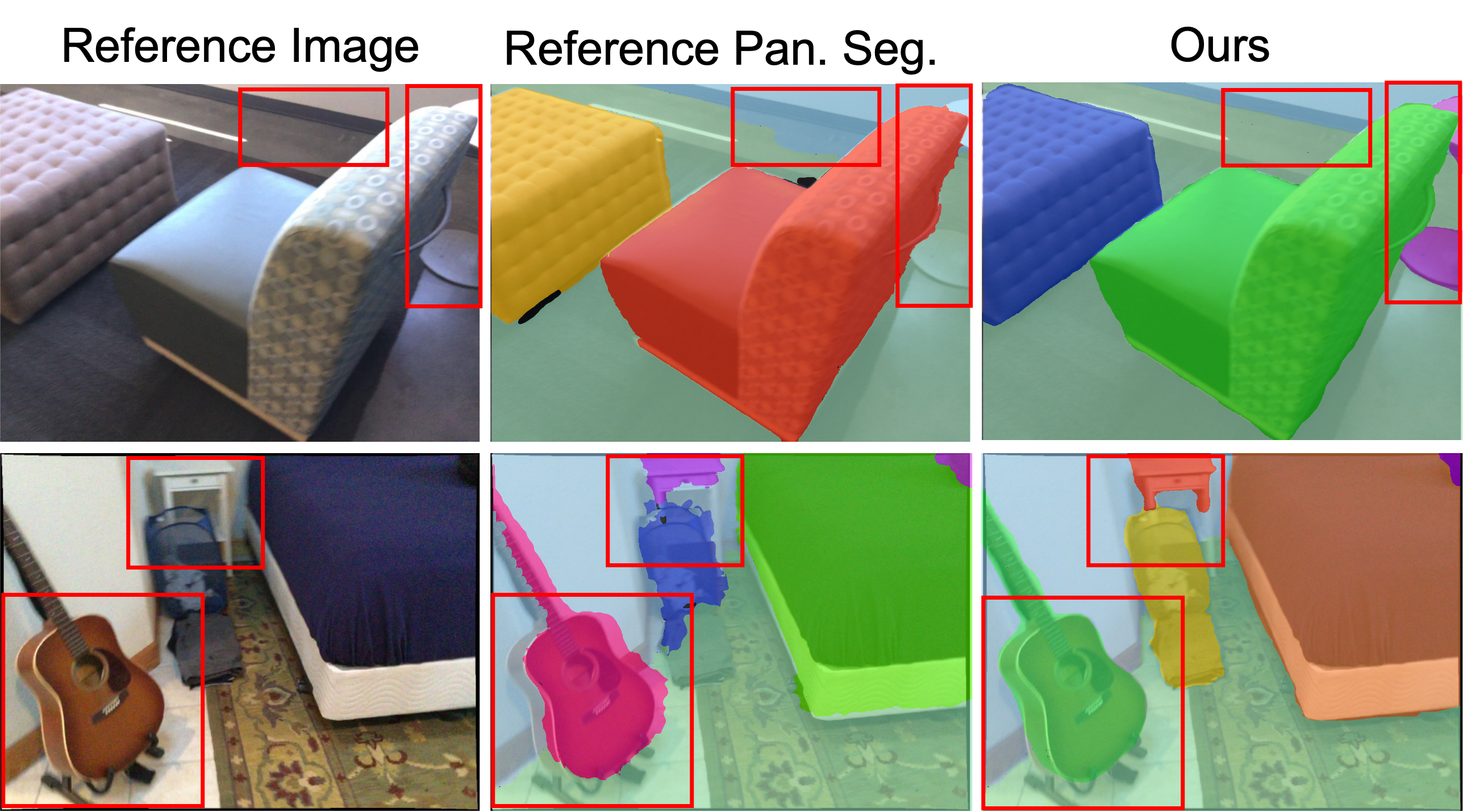}
   \caption{Comparison of ScanNet reference panoptic segmentation maps and our panoptic segmentation predictions overlaid on reference images.}
   \label{fig:sup_scannet_fix}
\end{figure}

In Figure~\ref{fig:sup_scannet_fix}, we visualize more examples on ScanNet comparing our panoptic predictions with reference annotations from the dataset. It can be seen that our representation learned from 2D supervision contains rich information and can produce more accurate segmentation results than reference maps in some cases. Our method produces clearer boundaries, fewer holes and discovers missing objects in the reference results, thanks to its ability to fuse segmentations from multiple views with a 3D sparsity prior from the structure of the representation.


{\small
\bibliographystyle{ieee_fullname}
\bibliography{egbib}
}

\end{document}